\documentclass[11pt,a4paper]{article}
\usepackage[hyperref]{acl2021}
\usepackage{times}
\usepackage{latexsym}
\usepackage{setspace}
\usepackage{color}
\usepackage{array}

\newcolumntype{"}{@{\hskip\tabcolsep\vrule width 1pt\hskip\tabcolsep}}
\usepackage[T1]{fontenc}

\usepackage[utf8]{inputenc}
\usepackage{graphicx}
\usepackage{microtype}
\usepackage{booktabs}
\usepackage{amsmath,amsthm,amsfonts,amssymb,bm}
\usepackage{mathrsfs}
\usepackage{amsmath}
\usepackage{algorithm}  
\usepackage{algorithmicx}  
\usepackage{algpseudocode}  
\usepackage{booktabs}
\def\aclpaperid{420} 
\usepackage{enumitem}
\usepackage{amsthm}
\usepackage{amssymb}
\newcommand{\tabincell}[2]{\begin{tabular}{@{}#1@{}}#2\end{tabular}}
\theoremstyle{definition}
\usepackage{multirow}
\usepackage{verbatim}
\usepackage{etoolbox}
\usepackage{hyperref}
\usepackage{subfigure}
\makeatletter
\def\UrlAlphabet{%
      \do\a\do\b\do\c\do\d\do\e\do\f\do\g\do\h\do\i\do\j%
      \do\k\do\l\do\m\do\n\do\o\do\p\do\q\do\r\do\s\do\t%
      \do\u\do\v\do\w\do\x\do\y\do\z\do\A\do\B\do\C\do\D%
      \do\E\do\F\do\G\do\H\do\I\do\J\do\K\do\L\do\M\do\N%
      \do\O\do\P\do\Q\do\R\do\S\do\T\do\U\do\V\do\W\do\X%
      \do\Y\do\Z}
\def\UrlDigits{\do\1\do\2\do\3\do\4\do\5\do\6\do\7\do\8\do\9\do\0}
\g@addto@macro{\UrlBreaks}{\UrlOrds}
\g@addto@macro{\UrlBreaks}{\UrlAlphabet}
\g@addto@macro{\UrlBreaks}{\UrlDigits}
\usepackage{microtype}

\aclfinalcopy 

\newcommand{\thickhline}{
    \noalign {\ifnum 0=`}\fi \hrule height 1pt
    \futurelet \reserved@a \@xhline
}
\setitemize[1]{leftmargin=10pt,itemsep=0.5pt,partopsep=0pt,parsep=0.5pt,topsep=0pt}
\title{Guiding the Growth: Difficulty-Controllable Question Generation through Step-by-Step Rewriting}

\author{Yi Cheng$^{1}$, Siyao Li$^{2}$, Bang Liu$^{3}$\thanks{~~\footnotesize{Corresponding author.}}, Ruihui Zhao$^{1}$, Sujian Li$^{4}$, Chenghua Lin$^{5}$, Yefeng Zheng$^{1}$\\
$^1$Tencent Jarvis Lab, China \quad
$^2$LTI, Carnegie Mellon University\\
$^3$RALI \& Mila, Université de Montréal\quad
$^4$Peking University\quad
$^5$The University of Sheffield\\
 \tt \{yicheng, zacharyzhao, yefengzheng\}@tencent.com, \\
 \tt siyaol@andrew.cmu.edu, bang.liu@umontreal.ca, \\
 \tt lisujian@pku.edu.cn, c.lin@shef.ac.uk 
}

\date{}

\begin{document}
\maketitle
\begin{abstract}

This paper explores the task of Difficulty-Controllable Question Generation (DCQG), which aims at generating questions with required difficulty levels. 
Previous research on this task mainly defines the difficulty of a question as whether it can be correctly answered by a Question Answering (QA) system, lacking interpretability and controllability. 
In our work, we redefine question difficulty as the number of inference steps required to answer it and argue that Question Generation (QG) systems should have stronger control over the logic of generated questions. 
To this end, we propose a novel framework that progressively increases question difficulty through step-by-step rewriting under the guidance of an extracted reasoning chain. A dataset is automatically constructed 
to facilitate the research, on which extensive experiments are conducted to test the performance of our method. 

\end{abstract}

\section{Introduction}
The task of Difficulty-Controllable Question Generation (DCQG) aims at generating questions with required difficulty levels and has recently attracted researchers' attention due to its wide application, such as facilitating certain curriculum-learning-based methods for QA systems \cite{sachan2016curriculumQA} and designing exams of various difficulty levels for educational purpose \cite{Ghader2020educationQG}. 

Compared to previous QG works which control the interrogative word \cite{kang2018qType, Kang2019qType} or the context of a question \cite{liu2020asking,liu2019learning}, few works have been conducted on difficulty control, as it is hard to formally define the difficulty of a question.
To the best of our knowledge,  \citet{gao2019easyHardControllable} is the only previous work of DCQG for free text, and defines question difficulty as whether a QA model can correctly answer it. This definition gives only two difficulty levels and is mainly empirically driven, lacking interpretability for what difficulty is and how difficulty varies.

\begin{figure}[t]
\centering
\vspace{1mm}
\includegraphics[width=\linewidth]{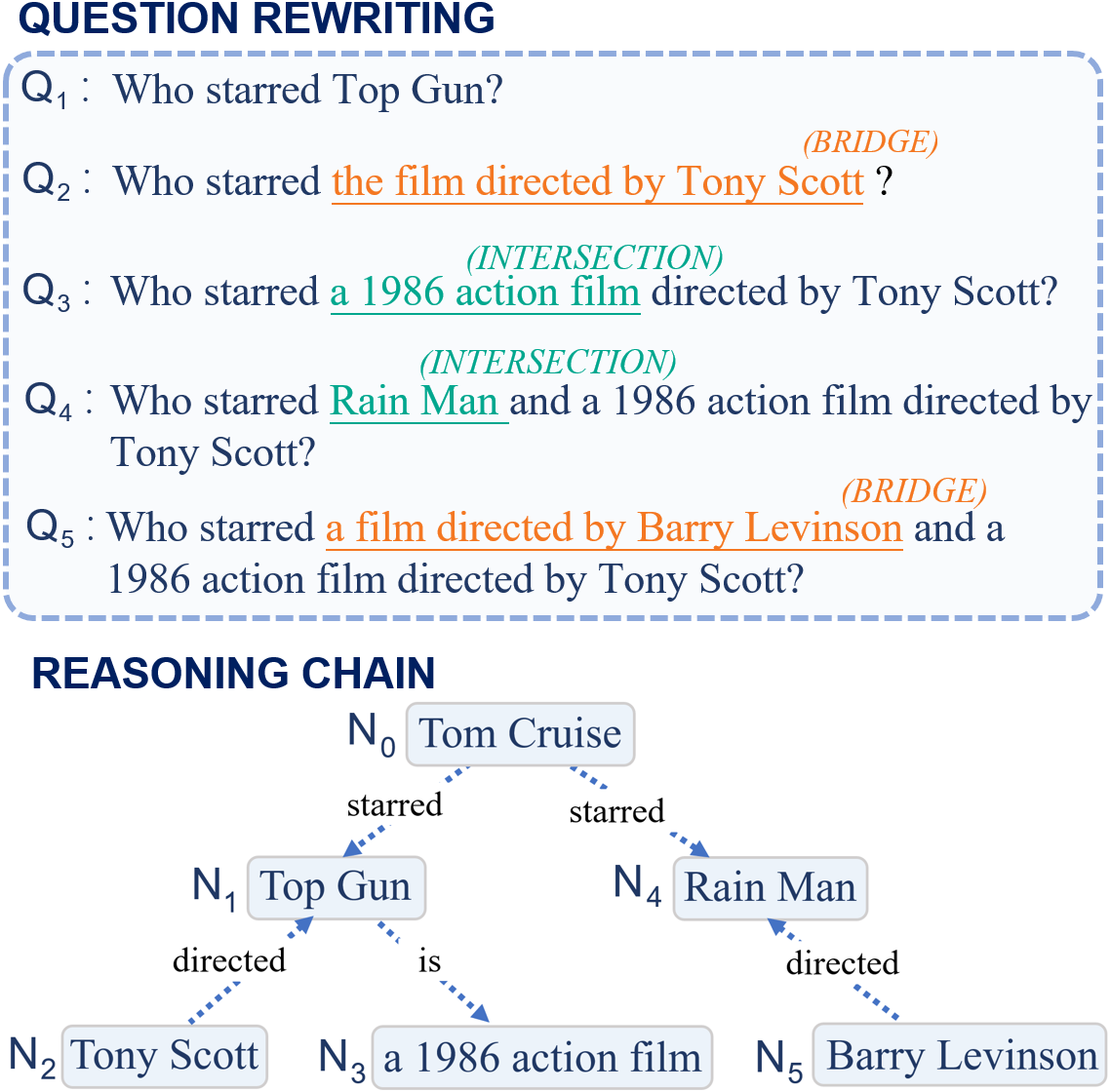}
\caption{An example of generating a complex question through step-by-step rewriting based on the reasoning chain extracted from a constructed context graph. }
\vspace{2mm}
\label{Figure:RewritingExp}
\end{figure}

In this work, we redefine the difficulty level of a question as \emph{the number of inference steps required to answer it}, which reflects the requirements on reasoning and cognitive abilities \cite{pan2019QGsurvey}. 
Existing QA systems perform substantially worse in answering multi-hop questions than single-hop ones \cite{Yang2018HotpotQA}, 
also supporting the soundness of using reasoning hops to define difficulty.


To achieve DCQG with the above definition, a QG model should have strong control over the logic and reasoning complexity of generated questions. Graph-based methods are well suited for such logic modelling \cite{pearl1985graphlogic, Yuyu2020graphlogic}. In previous QG researches, \citet{yu2020multihopWWW} and \citet{pan2020DQG} implemented graph-to-sequence frameworks to distill the inner structure of the context, but they mainly used graphs to enhance document representations, rather than to control the reasoning complexity of questions. 

In this paper, we propose a highly-controllable QG framework that progressively increases difficulties of the generated questions through step-by-step rewriting.
Specifically, we first transform a given raw text into a context graph, from which 
we sample the answer and the reasoning chain for the generated question. 
Then, we design a question generator and a question rewriter to generate an initial simple question and step-by-step rewrite it into more complex ones. 
As shown in Fig.~\ref{Figure:RewritingExp}, ``Tom Cruise'' is the selected answer, and 
$Q_1$ is the initial question, which is then adapted into $Q_2$ by adding one more inference step (i.e. $N_1$$\leftarrow$$N_2$) in the reasoning chain. That is, it requires to infer ``Top Gun'' is ``the film directed by Tony Scott'' before answering $Q_1$. 
Similarly, we can further increase its difficulty level and step-by-step extend it into more difficult questions (i.e., $Q_3$, $Q_4$ and $Q_5$). 

To train our DCQG framework, we design effective strategies to automatically construct the training data from existing QA datasets instead of building one from scratch with intensive human efforts. 
Specifically, we utilize HotpotQA \cite{Yang2018HotpotQA}, a QA dataset where most questions require two inference steps to answer and can be decomposed into two 1-hop questions.
Thus, we get the dataset that contains 2-hop questions and their corresponding 1-hop reasoning steps.
Having learned how to rewrite 1-hop questions into 2-hop ones with this dataset, our framework can easily extend to the generation of ($n$+1)-hop questions from $n$-hop ones only with a small amount of corresponding data, 
because the rewriting operation follows rather certain patterns regardless of the exact value of $n$, as shown in  Fig.~\ref{Figure:RewritingExp}.  

Extensive evaluations show that our method can controllably generate questions with required difficulty, and keep competitive question quality at the same time, compared with a set of strong baselines. 

In summary, our contributions are as follows:
\begin{itemize}
\item To the best of our knowledge, this is the first work of difficulty-controllable question generation, with question difficulty defined as the inference steps to answer it; 
\item We propose a novel framework that achieves DCQG through step-by-step rewriting under the guidance of an extracted reasoning chain; 
\item We build a dataset that can facilitate training of rewriting questions into more complex ones, paired with constructed context graphs and the underlying reasoning chain of the question. 
\end{itemize}

\section{Related Work}
\paragraph{Deep Question Generation} 
Most of the previous QG researches \cite{zhou2017NLG++,pan2019QGsurvey,liu2020asking} mainly focused on generating single-hop questions like the ones in SQuAD \cite{rajpurkar2016squad}. 
In the hope that AI systems could provoke more in-depth interaction with humans, deep question generation aims at generating questions that require deep reasoning. Many recent works attempted to conquer this task with graph-based neural architectures. \citet{Talmor2018WebComplexQ} and \citet{kumar2019multihopKG} generated complex questions based on knowledge graphs, but their methods could not be directly applied to QG for free text, which lacks clear logical structures. 
In sequential question generation, 
\citet{chai2020dualGraph} used a dual-graph interaction to better capture context dependency. However, they considered all the tokens as nodes, which led to a very complex graph. \citet{yu2020multihopWWW} tried to generate multi-hop questions from free text with the help of entity graphs constructed by external tools. Our work shares a similar setting with \citet{yu2020multihopWWW}, and we further explore the problem of how to generate deep questions in a more controllable paradigm.

\paragraph{Difficulty-Controllable Question Generation} 
DCQG is a relatively new task. \citet{gao2019easyHardControllable} classified questions as easy or hard according to whether they could be correctly answered by a BERT-based QA model, and controlled the question difficulty by modifying the hidden states before decoding. Another research on QG for knowledge graphs \cite{kumar2019multihopKG} estimated the question difficulty based on popularity of the named entity. They manipulated the generation process by incorporating the difficulty level into the input embedding of the Transformer-based decoder. 
In our work, we control the question difficulty based on the number of its reasoning hops, which is more explainable. 

\begin{figure*}[t]
\centering
\includegraphics[width=\linewidth]{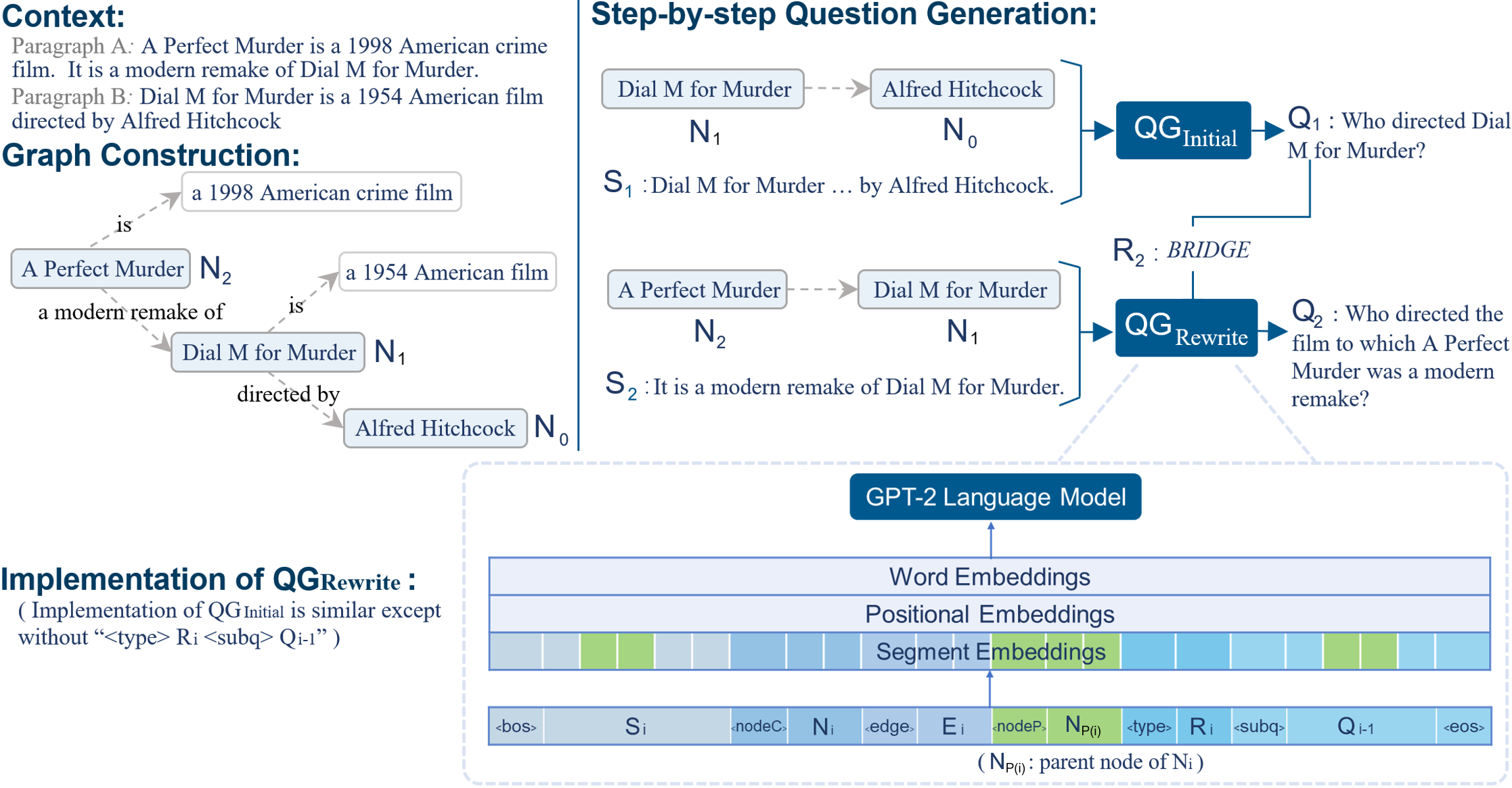}
\caption{An overview of our proposed framework. The selected reasoning chain is marked as light blue nodes.}
\label{Figure:Framework}
\end{figure*}

\paragraph{Question Rewriting} It is another emerging trend in the recent researches, demonstrating benefits to both QG and QA tasks.
With rewriting, QG models produced more complex questions by incorporating more context information into simple questions \cite{elgohary2019can,vakulenko2020question}, and QA pipelines could also decompose the original complex question into multiple shorter questions to improve model performance \cite{min2019decomposeQA,Khot2020decomposeQA}.

\section{Method}

Given input context text $\mathcal{C}$ and a specific difficulty level $d$, 
our objective is to generate a \emph{(question, answer)} pair $({\mathcal{Q}},{\mathcal{A}})$, where ${\mathcal{A}}$ is a sub-span of $\mathcal{C}$ and $\mathcal{Q}$ requires $d$-hop reasoning to answer.
Fig.~\ref{Figure:Framework} and Algorithm~\ref{alg:DCQGFramework} give an overview of our proposed framework.
First, we construct a context graph $\mathcal{G}_{CG}$ corresponding to the given context, from which a subgraph $\mathcal{G}_T$ is selected to serve as the reasoning chain of the generated question.
Next, with the reasoning chain and other contextual information as input, a question generator 
(QG$_{\rm Initial}$) produces an initial simple question $\mathcal{Q}_1$.
Then, $\mathcal{Q}_1$ is fed to a question rewriting module (QG$_{\rm Rewrite}$), which iteratively rewrites it into a more complex question $\mathcal{Q}_i$ $(i=2,3,\dots,d)$. 
In what follows, we will introduce the whole generation process in more details.

\paragraph{Context Graph Construction}
We follow the method proposed by ~\citet{fan2019localGraph} to build the context graph $\mathcal{G}_{CG}$. Specifically, we first apply open information extraction \cite{Stanovsky2018OpenIE} to extract $\langle subject, relation, object\rangle$ triples from context sentences. Each triple is then transformed into two nodes connected with a directed edge, like \emph{A Perfect Murder} $\stackrel{is}{\longrightarrow}$ \emph{a 1998 American crime film} in Fig.~\ref{Figure:Framework}. The two nodes respectively represent the subject and object, and the edge describes their relation.
Coreference resolution \cite{Lee2017AllenNlpCoreferenceResolution} is applied to merge nodes referring to the same entity.
For instance, \emph{A Perfect Murder} is merged with \emph{It} in Fig.~\ref{Figure:Framework}. 

\paragraph{Reasoning Chain Selection}
With the context graph constructed, we sample a connected subgraph $\mathcal{G}_T$ consisting of $d+1$ nodes from it to serve as the reasoning chain of the generated question.
A node $\mathcal{N}_{0}$ is first sampled as the answer of the question, if it is, or linked with, a named entity that has more than one node degree.
Next, we extract from $\mathcal{G}_{CG}$ a maximum spanning tree $\mathcal{G}_L$, with $\mathcal{N}_{0}$ as its root node, e.g., the tree structure shown in Fig.~\ref{Figure:RewritingExp}. $\mathcal{G}_{CG}$ is temporarily considered as an undirected graph at this step.
We then prune $\mathcal{G}_L$ into $\mathcal{G}_T$ to keep only $d+1$ nodes.
During pruning, we consider the sentence position where each node is extracted in order to make the reasoning chain relevant to more context.
In the following, we will denote a node in $\mathcal{G}_T$ as $\mathcal{N}_i$ $(i=0,1,\dots,d)$, where each node is subscripted by \emph{preorder traversal} of $\mathcal{G}_T$, and $\mathcal{N}_{P(i)}$ as the parent of $\mathcal{N}_i$.

\begin{algorithm}[t]
\caption{Procedure of Our DCQG Framework}
\label{alg:DCQGFramework}
\small
\setstretch{1.2}
    \begin{algorithmic}[1]
    \Require context $\mathcal{C}$, difficulty level $d$
    \Ensure  $({\mathcal{Q}}, {\mathcal{A}})$
        \State $\mathcal{G}_{CG} \leftarrow \mathbf{BuildCG}(\mathcal{C})$
        \State $\mathcal{N}_0 \leftarrow \mathbf{SampleAnswerNode}(\mathcal{G}_{CG})$
        \State $\mathcal{G}_L \leftarrow \mathbf{MaxTree}(\mathcal{G}_{CG}, \mathcal{N}_0)$
        \State $\mathcal{G}_T \leftarrow \mathbf{Prune}(\mathcal{G}_L, d)$
        \For{$\mathcal{N}_i$ \textbf{in} $\mathbf{PreorderTraversal}(\mathcal{G}_T)
        $}
            \State \textbf{if} $i=0$ \textbf{then} continue
            \State $\mathcal{N}_{P(i)} = \mathbf{Parent}(\mathcal{N}_i)$
            \State $\mathcal{S}_i = \mathbf{ContextSentence}(\mathcal{C},\mathcal{N}_i,\mathcal{N}_{P(i)})$
            \State $\mathcal{R}_i \leftarrow \left\{
            \begin{array}{ll}
                \text{\emph{Bridge}}   & \text{if } {\mathcal{N}_i \text{=} \mathbf{FirstChild}(\mathcal{N}_{P(i)})}\\
               \text{\emph{Intersection}}  & \text{else}\\
            \end{array} \right.$
            \State $\mathcal{Q}_i \leftarrow \left\{
            \begin{array}{l}
                \mathbf{QG}_{Initial}(\mathcal{N}_i,\mathcal{N}_{P(i)}, \mathcal{S}_i)   \text{\ \quad\qquad if } {i=1}\\
                \mathbf{QG}_{Rewrite}(\mathcal{Q}_{i-1}, \mathcal{N}_i,\mathcal{N}_{P(i)}, \mathcal{S}_i,\mathcal{R}_i )  \text{ else}\\
            \end{array} \right.$
        \EndFor
        \State return $(\mathcal{Q}_d, \mathcal{N}_0)$
    \end{algorithmic}
\end{algorithm}

\paragraph{Step-by-step Question Generation}
Our step-by-step QG process is described at lines 5-11 in Algorithm~\ref{alg:DCQGFramework}.
The following notations are defined for clearer illustration:
\begin{itemize}
    \item ${\mathcal{Q}}_i$ $(i=1,2,\dots,d)$ represents the question generated at each step, where ${\mathcal{Q}}_d$ is the final question ${\mathcal{Q}}$, and 
    ${\mathcal{Q}}_{i+1}$ is rewritten from ${\mathcal{Q}}_{i}$ by adding one more hop of reasoning. 
    \item $\mathcal{S}_{i}$ represents the context sentence from which we extract the triple $\mathcal{N}_i \rightarrow \mathcal{N}_{P(i)}$. 
    \item $\mathcal{R}_i$ is the rewriting type of $\mathcal{Q}_i$ $(i=2,3,\dots,d)$. Specifically, we consider two types of rewriting patterns in this work: \emph{Bridge} and \emph{Intersection}. 
    As shown in Fig.~\ref{Figure:RewritingExp}, \emph{Bridge}-style rewriting replaces an entity with a modified clause, 
    while \emph{Intersection} adds another restriction to an existing entity in the question. These two types can be distinguished by whether $\mathcal{N}_i$ is the first child of its parent node, i.e., whether its parent node has already been rewritten once in \emph{Bridge} style.
\end{itemize}

To generate the final question with the required difficulty level $d$, we first use a question generator QG$_{\rm Initial}$ to generate an initial simple question based on $\mathcal{N}_1$, $\mathcal{N}_0$, and the corresponding context sentence $\mathcal{S}_1$.
Then, we repeatedly (for $d-1$ times) use QG$_{\rm Rewrite}$ to rewrite question $\mathcal{Q}_{i-1}$ into a more complex one $\mathcal{Q}_i$, based on node $\mathcal{N}_i$ and its parent node $\mathcal{N}_{P(i)}$, context sentence $\mathcal{S}_i$, and the rewriting type $\mathcal{R}_i$ $(i=2,3,\dots,d)$.
Formally, the generation process of QG$_{\rm Initial}$ and the rewriting process of QG$_{\rm Rewrite}$ can be defined as:

$$
{\mathcal{Q}}_1 = \mathop{\arg\max}_{\bar{\mathcal{Q}}_1} P(\bar{\mathcal{Q}}_1 | \mathcal{N}_1,\mathcal{N}_0, \mathcal{S}_1)$$
$$
{\mathcal{Q}}_i = \mathop{\arg\max}_{\bar{\mathcal{Q}}_i} P(\bar{\mathcal{Q}}_i | \mathcal{Q}_{i-1}, \mathcal{N}_i,\mathcal{N}_{P(i)}, \mathcal{S}_i,\mathcal{R}_i )$$
where $i=2,3,\dots,d$.

In our implementation, both QG$_{\rm Initial}$ and QG$_{\rm Rewrite}$ are initialized with the pre-trained GPT2-small model \cite{radford2019GPT2}, and then fine-tuned on our constructed dataset (see Sec.~\ref{section:data}).
The encoder of QG$_{\rm Rewrite}$, as illustrated in Fig.~\ref{Figure:Framework}, is similar to ~\citet{liu2020asking}. 
If $\mathcal{N}_i$ points to $\mathcal{N}_{P(i)}$, then the input sequence is organized in the form of 
``{\small{$\left< \text{\emph{bos}}\right>$ $\mathcal{S}_i$ $\left< \text{\emph{nodeC}}\right>$ $\mathcal{N}_{i}$ 
$\left< \text{\emph{edge}}\right>$ $\mathcal{E}_i$ $\left< \text{\emph{nodeP}} \right>$ $\mathcal{N}_{P(i)}$ 
$\left< \text{\emph{type}} \right>$ $\mathcal{R}_{i}$ 
$\left< \text{\emph{subq}}\right>$ ${\mathcal{Q}}_{i-1}$ 
$\left< \text{\emph{eos}}\right>$}}”, 
where $\mathcal{E}_i$ is the edge from $\mathcal{N}_{i}$ to $\mathcal{N}_{P(i)}$. The positions of ``{\small{$\left< \text{\emph{nodeC}}\right>$ $\mathcal{N}_{i}$}}'' and ``{\small{$\left< \text{\emph{nodeP}} \right>$ $\mathcal{N}_{P(i)}$}}'' will be exchanged if $\mathcal{N}_{P(i)}$ points to $\mathcal{N}_i$.
As for QG$_{\rm Initial}$, its input is organized in the same way except without  ``{\small{$\left< \text{type} \right>$ $\mathcal{R}_{i}$ $\left< \text{subq}\right>$ ${\mathcal{Q}}_{i-1}$}}”.

The segment embedding layer is utilized to identify different segments.
For those parts in $\mathcal{S}_i$ and $\mathcal{Q}_{i-1}$ that are the same as, or refer to the same entity as  $\mathcal{N}_{P(i)}$,
we replace their segment embeddings with the one of $\mathcal{N}_{P(i)}$, considering that the parent node of $\mathcal{N}_i$ plays an important role in denoting what to ask about, or which part to rewrite, as shown in Fig.~\ref{Figure:RewritingExp}.


\begin{algorithm}[t]
\setstretch{1.2}
\caption{Procedure of Data Construction}
\label{alg:DataConstruction}
\small
    \begin{algorithmic}[1]
    \Require context $\mathcal{C} = \{\mathcal{P}_1, \mathcal{P}_2\}$, 
    QA pair $({\mathcal{Q}}_2, {\mathcal{A}}_2)$, 
    supporting facts $\mathcal{F}$
    \Ensure  $\mathcal{R}_1, ({\mathcal{Q}}_1, {\mathcal{A}}_1),\mathcal{S}_1, \mathcal{S}_2, \{\mathcal{N}_0, \mathcal{E}_1, \mathcal{N}_1,  \mathcal{E}_2, \mathcal{N}_2\}$
        \State $\mathcal{R}_1 \leftarrow \mathbf{TypeClassify}({\mathcal{Q}}_2)$
        \State \textbf{if} $\mathcal{R}_1 \notin \{$\emph{Bridge}, \emph{Intersection}$\}$ \textbf{then} return
        \State $subq_1,subq_2 \leftarrow \mathbf{DecompQ}({\mathcal{Q}}_2)$
        \State $suba_1,suba_2 \leftarrow \mathbf{QA}(subq_1), \mathbf{QA}(subq_2)$
        \State ${\mathcal{Q}}_1, {\mathcal{A}}_1 \leftarrow \left\{
            \begin{array}{ll}
                subq_2, suba_2  & \text{if } {\mathcal{A}_2 = suba_2}\\
                subq_1, suba_1  & \text{else}\\
            \end{array} \right.$
        \State $\mathcal{S}_1, \mathcal{S}_2 \leftarrow \left\{
            \begin{array}{ll}
                \mathcal{F} \cap \mathcal{P}_1, \mathcal{F} \cap \mathcal{P}_2 & \text{if } {\mathcal{Q}}_1 \text{ concerns } \mathcal{P}_1\\
                \mathcal{F} \cap \mathcal{P}_2, \mathcal{F} \cap \mathcal{P}_1 & \text{else}\\
            \end{array} \right.$
        \State $\mathcal{N}_2 \leftarrow \mathbf{FindNode}(\mathcal{A}_2)$
        \State $\mathcal{N}_0, \mathcal{E}_1, \mathcal{N}_1, \mathcal{E}_2 \leftarrow \mathbf{Match}(subq_1, subq_2)$
    \end{algorithmic}
\end{algorithm}

\section{Automatic Dataset Construction}\label{section:data}
Manually constructing a new dataset for our task is difficult and costly. Instead, we propose to automatically build a dataset from existing QA datasets without extra human annotation.
In our work, the training data is constructed from HotpotQA \cite{Yang2018HotpotQA}, in which every context $\mathcal{C}$ consists of two paragraphs $\{\mathcal{P}_1, \mathcal{P}_2\}$, 
and most of the questions \emph{require two hops of reasoning, each concerning one paragraph}.
HotpotQA also annotates supporting facts $\mathcal{F}$, which are the part of the context most relevant to the question.
In addition to the information already available in HotpotQA, we also need the following information to train  QG$_{\rm Initial}$ and QG$_{\rm Rewrite}$:  
i) $(\mathcal{Q}_1,\mathcal{A}_1)$, the simple initial question and its answer, which are used to train QG$_{\rm Initial}$; 
ii) $\mathcal{R}_2$, the type of rewriting from ${\mathcal{Q}}_1$ to ${\mathcal{Q}}_2$; 
iii) $\{\mathcal{N}_0, \mathcal{N}_1, \mathcal{N}_2\}$, the reasoning chain of ${\mathcal{Q}}_2$; 
and iv) $\mathcal{S}_i$ $(i=1,2)$, the context sentences where we extract $\mathcal{N}_0$, $\mathcal{N}_1$ and $\mathcal{N}_2$.

Algorithm~\ref{alg:DataConstruction} describes our procedure to obtain the above information.
The construction process is facilitated with the help of a reasoning type classifier ($\mathbf{TypeClassify}$) and a question decomposer ($\mathbf{DecompQ}$), referring to \citet{min2019decomposeQA}. For each question in HotpotQA (i.e. ${\mathcal{Q}}_2$), we first distinguish its reasoning type, and filter out those that are not \emph{Bridge} and \emph{Intersection}. The reasoning type here corresponds to the rewriting type $\mathcal{R}_i$.
Then, $\mathbf{DecompQ}$ decomposes ${\mathcal{Q}}_2$ into two sub-questions, $subq_1$ and $subq_2$, based on span prediction and linguistic rules. For example, the $\mathcal{Q}_2$ in Fig.~\ref{Figure:Framework} will be decomposed into $subq_1$=\textit{``To which film A Perfect Murder was a modern remake?''}, and $subq_2$=\textit{``Who directed Dial M for Murder?''}.
After that, an off-the-shelf single-hop QA model \cite{min2019decomposeQA} is utilized to acquire the answer of the two sub-questions, which should be \textit{``Dial M for Murder''} and \textit{``Alfred Hitchcock''} in the example.

As for ${\mathcal{Q}}_1$, it is one of the sub-questions. When ${\mathcal{Q}}_2$ is of the \emph{Intersection} type, ${\mathcal{Q}}_1$ can be either $subq_1$ or $subq_2$. For the  \emph{Bridge} type, it is the sub-question that shares the same answer as ${\mathcal{A}}_2$. For the example above, ${\mathcal{Q}}_1$ is $subq_2$ because $suba_2 = {\mathcal{A}}_2$.
The context sentence $\mathcal{S}_i$ is supposed to provide supporting facts contained in the paragraph $\mathcal{F}$ that concerns ${\mathcal{Q}}_i$ $(i=1,2)$.
For the reasoning chain, it is selected from the local context graph by first locating $\mathcal{N}_2$ and then finding $\mathcal{N}_0, \mathcal{N}_1$ through text matching with the two sub-questions.

\section{Experiments} 
In the following experiments, we mainly evaluate the generation results of our proposed method when required to produce 1-hop and 2-hop questions, denoted as Ours$_{\rm 1\text{-}hop}$ and Ours$_{\rm 2\text{-}hop}$.
In Sec.~\ref{section:quality}, we compare our method with a set of strong baselines using both automatic and human evaluations on question quality. In Sec.~\ref{section:controlAnalysis}, we provide controllability analysis by manually evaluating their difficulty levels and testing the performance of QA systems in answering questions generated by different methods.
In Sec.~\ref{section:applyQA}, we test the effect of our generated QA pairs on the performance of a multi-hop QA model in a data augmentation setting. In Sec.~\ref{section:casestudy}, we further analyze the extensibility of our method, i.e., its potential in generating questions that require reasoning of more than two hops. 
Our code and constructed dataset have been made publicly available to facilitate future research.\footnote{$\,$
\url{https://tinyurl.com/19esunzz}}

\subsection{Experimental Setup}\label{section:expSetting}

\paragraph{Datasets} The constructed dataset described in Sec.~\ref{section:data} consists of 57,397/6,072/6,072 samples for training/validation/test. 
For context graph construction, we use the coreference resolution toolkit from AllenNLP 1.0.0 \cite{Lee2017AllenNlpCoreferenceResolution} and the open information extraction toolkit provided by the Plasticity developer API.\footnote{ $\,$\url{https://www.plasticity.ai/}} The question decomposer and the reasoning type classifier follow the implementations of~\citet{min2019decomposeQA}.  

\paragraph{Baselines} The following baselines are trained to generate the 2-hop questions in the datasets: 
\begin{itemize}
    \item \textbf{NQG++} \cite{zhou2017NLG++} is a seq2seq model based on bi-directional Gate Recurrent Unit (GRU), with features enriched by answer position and lexical information.
    \item \textbf{ASs2s} \cite{Kim2019ASs2s} is a seq2seq model based on Long Short-term Memory (LSTM), which separately encodes answer and context.
    \item \textbf{SRL-Graph} and \textbf{DP-Graph} \cite{pan2020DQG} are two state-of-the-art QG systems. They encode graph-level and document-level information with an attention-based Graph Neural Network (GNN) and a bi-directional GRU, respectively. 
    SRL-Graph constructs the semantic graph by semantic role labelling, and DP-Graph by dependency parsing.
    \item \textbf{GPT2$_{}$} 
    is a vanilla GPT2-based QG model. 
    Its input is the concatenation of context and sampled answer. The position where the answer appears in the context segment is denoted in the segment embedding layer. 
\end{itemize}

\paragraph{Implementation Details} The baseline models are trained to directly produce the 2-hop questions, while QG$_{\rm Initial}$ and QG$_{\rm Rewrite}$ are respectively trained to generate 1-hop questions and rewrite 1-hop ones into 2-hop. QG$_{\rm Initial}$, QG$_{\rm Rewrite}$, and GPT2$_{}$ are initialized with the GPT2-small model from the HuggingFace Transformer library \cite{Wolf2019huggingface}, and fine-tuned for 8, 10, and 7 epochs, respectively, with batch size of 16. We apply top-$p$ nucleus sampling with $p$ = 0.9 during decoding. AdamW \cite{Losh2017AdamW} is used as optimizer, with the initial learning rate set to be $6.25$$\times$$10^{-5}$ and adaptively decays during training. For DP-Graph, we use their released model and code to perform the experiment. For the other three baselines, we directly refer to the experiment results reported in \citet{pan2020DQG}. The performances of these baselines are compared under the same setting as in \citet{pan2020DQG}, where each context is abbreviated to only include the supporting facts and the part that overlaps with the question. More implementation details can be found in our code and the supplementary materials.

\subsection{Evaluation of Question Quality}\label{section:quality}

\begin{table}[t]

\small
	\begin{center}
		\begin{tabular}{ l  c  c  c  c } \specialrule{1.2pt}{1pt}{1pt}
        \small{\textbf{Model}} & \small{\textbf{BLEU3}} & \small{\textbf{BLEU4}} & \small{\textbf{METEOR}} & \small{\textbf{CIDEr}} \\ \specialrule{1.2pt}{0pt}{0.5pt}
        NQG++ & 15.41 & 11.50 & 16.96 & -\\
        ASs2s & 15.21 & 11.29 & 16.78 & -\\ 
        SRL-Graph & 19.66 & 15.03 & 19.73 & - \\
        DP-Graph & 19.87 & 15.23 & 20.10 &  {1.40} \\ 
        GPT2$_{}$ & 20.98 & \textbf{15.59} & \textbf{24.19}  & 1.46 \\ \specialrule{0.7pt}{0.5pt}{0.5pt}
        Ours$_{\rm 2\text{-}hop}$ & \textbf{21.07} & 15.26 & 19.99 & \textbf{1.48} \\ \specialrule{1.2pt}{0.5pt}{0pt}
		\end{tabular}
	\end{center}
	\caption{Automatic evaluation results of the baseline models and the 2-hop questions generated by our method (Ours$_{\rm 2\text{-}hop}$).}
\label{tbl:performance_comparision}
\end{table}

\paragraph{Automatic Evaluation}
The automatic evaluation metrics are BLEU3, BLEU4 \cite{Papineni2002BLEU}, METEOR \cite{Lavie2007Meteor}, and CIDEr \cite{vedantam2015cider}, which measure the similarity between the generation results and the reference questions in terms of $n$-grams.
As the four baselines are trained to generate 2-hop questions only, we only compare them with Ours$_{\rm 2\text{-}hop}$. 
As shown in Table~\ref{tbl:performance_comparision}, we can see that Ours$_{\rm 2\text{-}hop}$ and GPT2$_{}$ perform consistently better than the others. Though the performances of Ours$_{\rm 2\text{-}hop}$ and GPT2$_{}$ are close in terms of automatic metrics, we observe that the questions generated by Ours$_{\rm 2\text{-}hop}$ are usually more well-formed, concise and answerable, as illustrated in Table~\ref{tbl:example}. These advantages cannot be reflected through automatic evaluation. 

\paragraph{Human Evaluation}\label{section:HumanEvaluationQuality}
We randomly sample 200 questions respectively from DP-Graph, GPT2$_{}$, Ours$_{\rm 1\text{-}hop}$,  Ours$_{\rm 2\text{-}hop}$, as well as the reference 1-hop and 2-hop questions in the constructed dataset (Gold$_{\rm 1\text{-}hop}$, Gold$_{\rm 2\text{-}hop}$).
The questions are manually evaluated by eight human annotators, who are graduate students, majoring in English Literature, Computer Science, or Electronic Engineering. They voluntarily offer to help without being compensated in any form. Before annotation, they are informed of the detailed annotation instruction with clear scoring examples. 
The generated questions are evaluated in the following four dimensions: 
\begin{itemize}
    \item \textbf{Well-formed}: It checks whether a question is semantically correct. Annotators are asked to mark a question as \emph{yes}, \emph{acceptable}, or \emph{no}. \emph{Acceptable} is selected if the question is not grammatically correct, but its meaning is still inferrable. 
    \item \textbf{Concise}: It checks whether the QG models are overfitted, generating questions with redundant modifiers. The question is marked as \emph{yes} if no single word can be deleted, \emph{acceptable} if it is a little lengthy but still in a natural way, and \emph{no} if it is abnormally verbose. 
    \item \textbf{Answerable}: It checks whether a question is answerable according to the given context. The anonnotion is either \emph{yes} or \emph{no}.  
    \item \textbf{Answer Matching}: It checks whether the given answer is the correct answer to the question. The anonnotion is either \emph{yes} or \emph{no}. 
\end{itemize}

\begin{table}[t]

\footnotesize
	\begin{center}
        \begin{tabular}{m{3.4cm}m{3.55cm}}\specialrule{1.2pt}{0pt}{1pt}
\multicolumn{1}{c}{\textbf{Ours$_{\rm 2\text{-}hop}$}}                                                                                                                    &  \multicolumn{1}{c}{\textbf{GPT2$_{}$}}                                \\ \specialrule{1.2pt}{.5pt}{.5pt}
 When was the first theatre director of African descent born?    & When was the first theatre director of African descent to establish a national touring company in the UK born?    \\ \specialrule{0.7pt}{.5pt}{.5pt}
What play by Carrie Hamilton was run at the Goodman Theatre in 2002?  &  What play by Carrie Hamilton and Carol Burnett ran at the Goodman Theatre and on Broadway in 2002?\\ \specialrule{0.7pt}{.5pt}{.5pt}
What was the review score for the album that has been reissued twice?  &  What was the review of the album that includes previously unreleased tracks by Guetta from its first major international release? \\
        \specialrule{1.2pt}{.5pt}{0pt}        
        \end{tabular}
	\end{center}
\caption{Examples of generation results from Ours$_{\rm 2\text{-}hop}$ and GPT2$_{}$}
\label{tbl:example}
\end{table}

\begin{table*}[t]
\footnotesize
\begin{center}
\begin{tabular}{cc|ccc|ccc|cc|cc}
\specialrule{1.2pt}{1pt}{1pt}
\multirow{2}{*}{\bf \tabincell{c}{Difficulty\\Level}} & \multirow{2}{*}{\textbf{Model}} & \multicolumn{3}{c|}{\textbf {Well-formed}} & \multicolumn{3}{c|}{\textbf {Concise}} & \multicolumn{2}{c|}{\textbf {Answerable}} & \multicolumn{2}{c}{\textbf{Answer Matching}}      \\ \cline{3-12} 
\multicolumn{1}{c}{}  &                   & Yes              & Acceptable   & No    & Yes              & Acceptable    & No  & Yes                      & No             & Yes                                & No                     \\ \specialrule{1.2pt}{0pt}{0.5pt}
\multicolumn{1}{c}{\multirow{4}{*}{\textbf{2-hop}}} & DP-Graph                        & 28\%             & 41\%             & 31\%  & 41\%             & 53\%          & 6\% & 49\%                     & 51\%           & 39\%                               & 61\%                \\
 & GPT2$_{}$                         & 57\%             & 34\%             & 9\%    & 47\%             & 50\%          & 3\%   & 69\%                     & 31\%           & 66\%                               & 34\%                \\ 
 & Ours$_{\rm 2\text{-}hop}$                         & \textbf{74\%}    & 19\%             & 7\% & \textbf{67\%}    & 30\%          & 3\%  & \textbf{78\%}            & 22\%           & \textbf{69\%} & 31\%              \\  
 & Gold$_{\rm 2\text{-}hop}$                   & 72\%    & 22\%             & 6\%   &    56\%    & 40\%          & 4\%       & 92\%                      & 8\%      & 87\%            & 13\%          \\ \specialrule{0.8pt}{0.5pt}{0.5pt}
\multicolumn{1}{c}{\multirow{2}{*}{\textbf{1-hop}}} & Ours$_{\rm 1\text{-}hop}$    & \textbf{46\%} & 46\% & 8\% & \textbf{65\%} & 25\% & 10\% & \textbf{81\%} & 19\% & \textbf{72\%} & 28\%  \\
 & Gold$_{\rm 1\text{-}hop}$  & 56\% & 39\%	& 5\% & 80\% & 16\% & 4\% & 84\% & 16\% & 79\% & 21\% \\
 \specialrule{1.2pt}{0.5pt}{0pt}
\end{tabular}
\end{center}
\caption{Human evaluation results of question quality. }
\label{Table:HumanEvaluation}
\end{table*}

The results are shown in Table ~\ref{Table:HumanEvaluation}. Overall, we can see that Ours$_{\rm 2\text{-}hop}$ performs consistently better than DP-Graph and GPT2$_{}$ across all metrics and comparable to the hand-crafted reference questions. 
Our method performs especially well in terms of \emph{concise}, even better than the reference questions. For reference, the average word number of the questions generated by DP-Graph, GPT2$_{}$, Ours$_{\rm 2\text{-}hop}$, and Gold$_{\rm 2\text{-}hop}$ are 19.32, 19.26, 17.18, 17.44, respectively. It demonstrates that the enriched graph information and our multi-stage rewriting mechanism indeed enhance the question structure and content. In comparison, we find that the questions generated by the two baselines tend to unreasonably pile too many modifiers and subordinate clauses. 
As for the 1-hop questions, Ours$_{\rm 1\text{-}hop}$ performs well in terms of \emph{answerable} and \emph{answer matching}, but not so competitive in terms of \emph{well-formed}, mainly due to the limitation of its training data. As the 1-hop reference questions (Gold$_{\rm 1\text{-}hop}$) are automatically decomposed from the hand-crafted 2-hop questions, a significant portion (44\%) of them have some grammatical errors, but most of them are still understandable despite that.

\begin{table}[t]

\small
	\begin{center}
\begin{tabular}{c|cccc}
\specialrule{1.2pt}{1pt}{1pt}
\multirow{2}{*}{\textbf{Model}} & \multicolumn{4}{c}{\textbf{Inference Steps}}     \\ \cline{2-5}
                                & 1-hop & 2-hop & 3-hop & \textgreater{}3-hop   \\ \specialrule{1pt}{0pt}{0.5pt}

DP-Graph            & 26.1\%     & 55.1\%     & 8.7\%      & 10.1\%   \\
GPT2$_{}$            & 23.3\%     & 57.1\%     & 13.2\%      & 6.4\%   \\
Ours$_{\rm 2\text{-}hop}$            & 4.3\%      & \textbf{67.7\%}      & 25.8\%      & 2.2\%                     \\ 
Ours$_{\rm 1\text{-}hop}$            & \textbf{70.7\%}     & 28.2\%      & 1.1\%      &  0.0\%                    \\
\specialrule{1.2pt}{0.5pt}{0pt}
\end{tabular}
	\end{center}

\caption{Human evaluation results of the number of inference steps required by the generated questions.}
\vspace{3mm}
\label{tbl:ControlAnalysisHuman}
\end{table}

\vspace{2mm}
\subsection{Controllability Analysis}\label{section:controlAnalysis}
\vspace{2mm}
\paragraph{Human Evaluation of Controllability}
For controllability analysis, we manually evaluate the numbers of inference steps involved in generated questions. 
DP-Graph and GPT2$_{}$ are also evaluated for comparison. The results are shown in Table~\ref{tbl:ControlAnalysisHuman}.
70.65\% of Ours$_{\rm 1\text{-}hop}$ require one step of inference and 67.74\% of Ours$_{\rm 2\text{-}hop}$ require two steps, proving that our framework can successfully control the number of inference steps of most generated questions. 
In comparison, DP-Graph and GPT2$_{}$ are not difficulty-aware and their generated questions are more scattered in difficulty levels.

\paragraph{Difficulty Assessment with QA Systems}
For further assessment of question difficulty, we test the performance of QA models in answering questions generated by different models. Specifically, we utilize two off-the-shelf QA models provided by the HuggingFace Transformer library \cite{Wolf2019huggingface}, which are respectively initialized with BERT \cite{delvin2019BERT} and RoBERTa \cite{yinhan2019roberta}, and then fine-tuned on SQuAD \cite{rajpurkar2016squad}. 
We select those generated questions that are ensured to be paired with correct answers by the human evaluation described in Sec.~\ref{section:HumanEvaluationQuality}, and test the performance of two QA models in answering them. The evaluation metrics include Exact Match (EM) and F1. 

The results are shown in Table~\ref{tbl:ControlAnalysisQA}. 
We can see that questions generated by  Ours$_{\rm 2\text{-}hop}$ are more difficult than Ours$_{\rm 1\text{-}hop}$ not only to humans (requiring more hops of reasoning), but also to the state-of-the-art QA models.
In comparison, with a more scattered mix of 1-hop and 2-hop questions, the performances on DP-Graph and GPT2$_{}$ are between Ours$_{\rm 1\text{-}hop}$ and Ours$_{\rm 2\text{-}hop}$. This result demonstrates that our method can controllably generate questions of different difficulty levels for QA systems and that inference steps can effectively model the question difficulty.

\vspace{2mm}
\subsection{Boosting Multi-hop QA Performance}\label{section:applyQA}
\vspace{2mm}
\begin{table}[t]
\small
\begin{center}
	\begin{tabular}{ c | c  c  | c  c} \specialrule{1.2pt}{1pt}{1pt}
    \multirow{2}{*}{\textbf{Test Set}}  & \multicolumn{2}{c|}{\textbf{BERT}} & \multicolumn{2}{c}{\textbf{RoBERTa}}    \\ \cline{2-5} 
    & \small{EM} & \small{F1} & \small{EM} & \small{F1}\\ \specialrule{1pt}{0pt}{0.5pt}
    DP-Graph & 0.436 & 0.615  & 0.552 & 0.678 \\
    GPT2$_{}$ & 0.419  & 0.581 & 0.669 & 0.772\\ \specialrule{0.7pt}{0.5pt}{0.5pt}
    Ours$_{\rm 2\text{-}hop}$ & 0.295 & 0.381 & 0.506 &  0.663 \\
    Ours$_{\rm 1\text{-}hop}$ & 0.618 & 0.737 & 0.882 & 0.937\\
    \specialrule{1.2pt}{0.5pt}{0pt}
	\end{tabular}
\end{center}
\caption{Performance of BERT- and RoBERTa-based QA models on different generated QA datasets.}
\vspace{3mm}
\label{tbl:ControlAnalysisQA}
\end{table}

\begin{figure}[!t]
	\centering
	\!\!\!\!\!
	\subfigure{
		\begin{minipage}[t]{0.46\linewidth}
			\centering
			\includegraphics[width=1.5in]{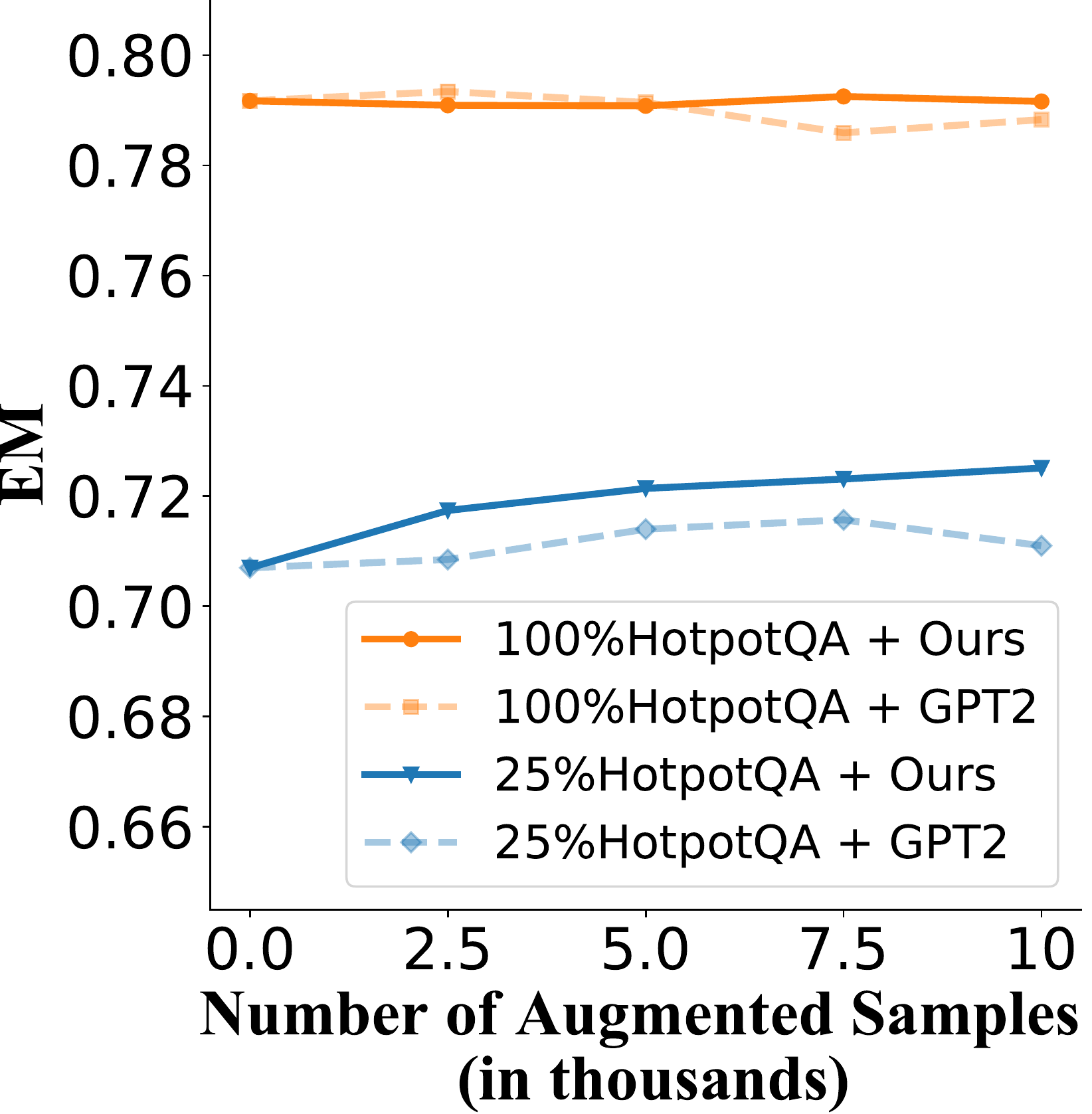}
		\end{minipage}
	}
	\subfigure{
		\begin{minipage}[t]{0.46\linewidth}
			\centering
			\includegraphics[width=1.5in]{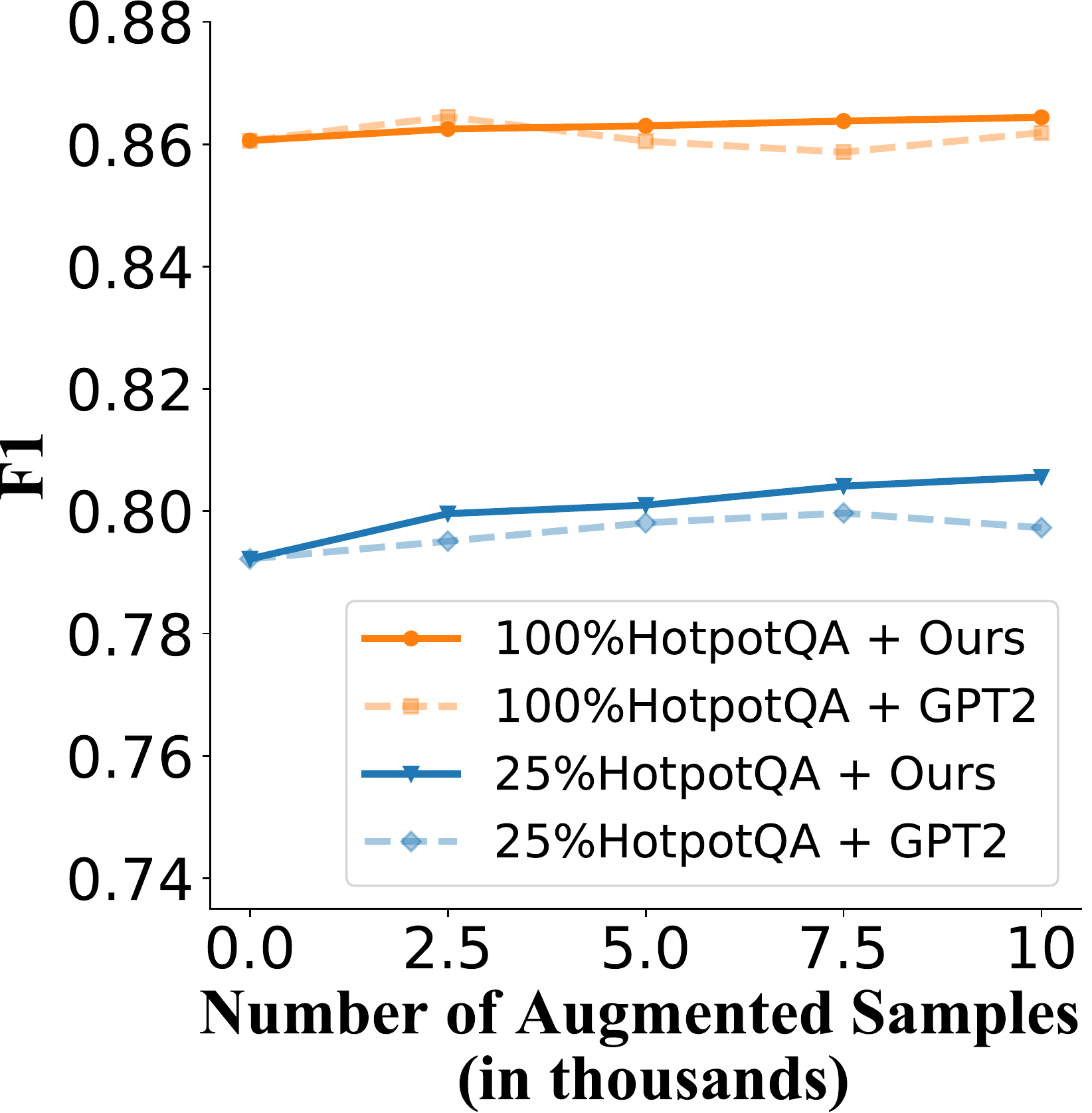}
		\end{minipage}
	}
	\caption{Performance of the DistilBERT-based QA system on HotpotQA, augmented with different quantities of generated data.}
	\label{Figure:QA_boost}
\end{figure}

We further evaluate whether the generated QA pairs can boost QA performance through data augmentation. Specifically, we heuristically sample the answers and reasoning chains from the context graphs in our constructed dataset to generate 150,305 two-hop questions. 
As a comparison, we utilize GPT2$_{}$ to generate the same amount of data with the same sampled answers and contextual sentences. 
Some low-quality questions are filtered out if their word counts are not between 6$\sim$30 (4.7\% for ours and 9.2\% for GPT2$_{}$), or the answers directly appear in the questions (2.7\% for ours and 2.4\% for GPT2$_{}$).
Finally, we randomly sample 100,000 QA pairs and augment the HotpotQA dataset with them. 

A DistilBERT-based \cite{Sanh2019DistilBERT} QA model is implemented. 
It takes as input the concatenation of context and question to predict the answer span. 
To speed up the experiment, we only consider those necessary supporting facts as the question answering context. 
During training, the original samples from HotpotQA are oversampled to ensure that they are at least 4 times as the generated data.
We use Adam \cite{Kingma2014Adam} as the optimizer, with the mini-batch size of 32.
The learning rate is initially set to $3$$\times$$10^{-5}$ and adaptively decays during training.
The configurations are the same in all the QA experiments, except that the training datasets are different combinations of HotpotQA and the generated data. 
The validation and test sets are the same as those in HotpotQA.

We test the impact of the generated data under both \emph{high-resource} (using the whole training set of HotpotQA) and  \emph{low-resource} settings (using only 25\% of the data randomly sampled from HotpotQA). Fig.~\ref{Figure:QA_boost} compares the QA performance, augmented with different quantities of the data generated by our method and by GPT2$_{}$, respectively.
We can see that under both settings, our method achieves better performance than GPT2$_{}$. 
Under the low-resource setting, performance boost achieved by our generated data is more significant and obviously better than that of GPT2$_{}$. The performance of the QA model steadily improves when the training dataset is augmented with more data. EM and F1 of the QA model are improved by 2.56\% and 1.69\%, 
respectively, when 100,000 samples of our generated data are utilized.


\vspace{4mm}
\subsection{More-hop Question Generation}\label{section:casestudy}
\vspace{2mm}

\begin{figure}[t]
    \centering
    \includegraphics[width=\linewidth]{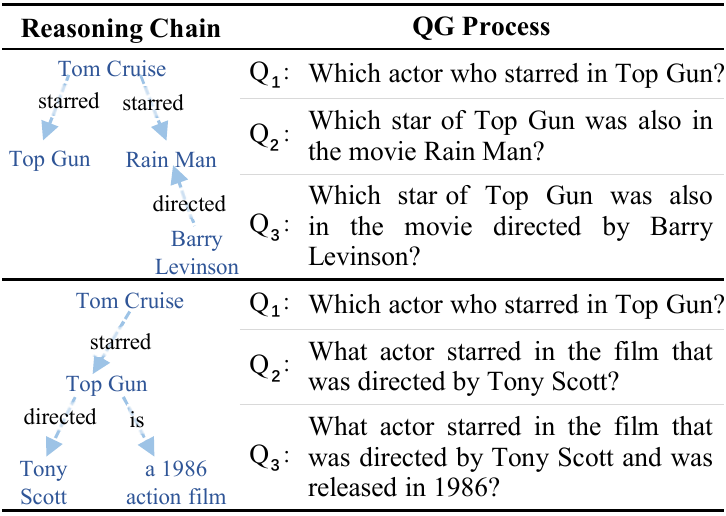}
    \caption{Two examples of generating three-hop questions based on the extracted reasoning chains.}
    \label{Figure:CaseStudy}
\end{figure}

To analyze the extensibility of our method, we experiment with the generation of questions that are more than 2-hop, by repeatedly using QG$_{\rm Rewrite}$ to increase question difficulty.
Fig.~\ref{Figure:CaseStudy} shows two examples of 3-hop question generation process. The two intermediate questions and the corresponding reasoning chains are also listed for reference. 

We can see that the intermediate questions, serving as springboards, are effectively used by QG$_{\rm Rewrite}$ to generate more complex questions. With the training data that only contains 1-hop and 2-hop questions, our framework is able to generate some high-quality 3-hop questions, demonstrating the extensibility of our framework. It can be expected that the performance of our model can be further strengthened if a small training set of 3-hop question data is available. 

Besides, it can also be observed that though the contexts and answers of these two questions are the same, two different questions with different underlying logic are generated, illustrating that the extracted reasoning chain effectively controls the question content.

However, when generating questions with more than 3 hops, we find that the question quality drastically declines. The semantic errors become more popular, and some content tend to be unreasonably repeated. 
It is probably because the input of QG$_{\rm Rewrite}$ has become too long to be precisely encoded by the GPT2-small model due to the growing length of the question. 
It will be our future work to explore how to effectively extend our method to more-hop question generation.

\section{Conclusion}
We explored the task of difficulty-controllable question generation, with question difficulty redefined as the inference steps required to answer it.  A step-by-step generation framework was proposed to accomplish this objective, with an input sampler to extract the reasoning chain, a question generator to produce a simple question, and a question rewriter to further adapt it into a more complex one. A dataset was automatically constructed based on HotpotQA to facilitate the research. Extensive evaluations demonstrated that our method can effectively control difficulty of the generated questions, and keep high question quality at the same time. 

\section*{Acknowledgments}
Thanks to Zijing Ou, Yafei Liu and Suyuchen Wang for their helpful comments on this paper.

\bibliographystyle{acl_natbib}
\bibliography{anthology,acl2021}


\end{document}